\documentclass{article}
\usepackage[preprint]{colm2026_conference}
\usepackage{microtype}
\usepackage{hyperref}
\usepackage{url}
\usepackage{booktabs}
\usepackage{lineno}
\definecolor{darkblue}{rgb}{0, 0, 0.5}
\hypersetup{colorlinks=true, citecolor=darkblue, linkcolor=darkblue, urlcolor=darkblue}
\usepackage{graphicx}
\usepackage{amsmath}
\usepackage{amssymb}
\usepackage{algorithm}
\usepackage{algorithmic}
\usepackage{multirow}
\usepackage{amsthm}
\usepackage{subcaption}
\usepackage{xcolor}
\usepackage{wrapfig}
\usepackage{colortbl}
\usepackage{enumitem}
\ifx\theorem\undefined
\newtheorem{theorem}{Theorem}

\newtheorem{proposition}{Proposition}

\title{Multi-Granularity Reasoning for Image Quality Assessment \\ via Attribute-Aware Reinforcement Learning to Rank}

\author{Xiangyong Chen, Xiaochuan Lin, Haoran Liu, Xuan Li, Yichen Su, Xiangwei Guo\\
Henan Polytechnic University\\
gxw@hpu.edu.cn}

\newcommand{\ours}{MG-IQA}

\begin{document}
\ifcolmsubmission
\linenumbers
\fi
\maketitle

\begin{abstract}
Recent advances in reasoning-induced image quality assessment (IQA) have demonstrated the power of reinforcement learning to rank (RL2R) for training vision-language models (VLMs) to assess perceptual quality. However, existing approaches operate at a single granularity, predicting only an overall quality score, while overlooking the multi-dimensional nature of human quality perception, which encompasses attributes such as sharpness, color fidelity, noise level, and compositional aesthetics. In this paper, we propose MG-IQA (Multi-Granularity IQA), a multi-granularity reasoning framework that extends RL2R to jointly assess overall image quality and fine-grained quality attributes within a single inference pass. Our approach introduces three key innovations: (1) an attribute-aware prompting strategy that elicits structured multi-attribute reasoning from VLMs; (2) a multi-dimensional Thurstone reward model that computes attribute-specific fidelity rewards for group relative policy optimization; and (3) a cross-domain alignment mechanism that enables stable joint training across synthetic distortion, authentic distortion, and AI-generated image datasets without perceptual scale re-alignment. Extensive experiments on eight IQA benchmarks demonstrate that MG-IQA consistently outperforms state-of-the-art methods in both overall quality prediction (average SRCC improvement of 2.1\%) and attribute-level assessment, while generating interpretable, human-aligned quality descriptions.
\end{abstract}

\section{Introduction}
\label{sec:intro}

Image quality assessment (IQA) is a fundamental problem in computer vision that aims to quantify the perceptual quality of images in alignment with human subjective perception~\citep{wang_2004_ssim}. With the proliferation of digital imaging applications, from smartphone photography~\citep{fang_2020_spaq} to AI-generated content~\citep{rombach_2022_ldm, saharia_2022_imagen, songbroad}, the demand for accurate, automated quality assessment has never been greater. No-reference IQA (NR-IQA), which evaluates image quality without access to a pristine reference, is particularly valuable for real-world deployment where reference images are typically unavailable~\citep{mittal_2012_brisque, mittal_2013_niqe}.

Traditional NR-IQA methods rely on handcrafted features derived from natural scene statistics~\citep{mittal_2012_brisque, mittal_2013_niqe}, while modern deep learning approaches employ discriminative models trained to regress mean opinion scores (MOS)~\citep{ke_2021_musiq, yang_2022_maniqa, zhang_2021_unique}. More recently, vision-language models (VLMs) have emerged as powerful backbones for IQA, leveraging their rich visual-linguistic representations to assess quality~\citep{wu_2023_q_align, zhang_2023_liqe, you_2025_deqa_score, wang_2022_clip_iqa}. A notable breakthrough is VisualQuality-R1~\citep{wu_2025_visualquality_r1}, which introduced reinforcement learning to rank (RL2R) by combining group relative policy optimization (GRPO)~\citep{shao_2024_grpo} with the Thurstone model~\citep{thurstone_1927_law} to train VLMs for quality-aware reasoning. Concurrent work Q-Insight~\citep{li_2025_q_insight} similarly applies GRPO for IQA, confirming the effectiveness of reasoning-induced quality assessment.

Despite these advances, current reasoning-induced IQA methods suffer from a fundamental limitation: they operate at a \emph{single granularity}, predicting only one holistic quality score per image. In practice, however, human quality perception is inherently \emph{multi-dimensional}, a viewer simultaneously evaluates sharpness, color accuracy, noise level, compositional aesthetics, and other attributes when judging image quality. This mismatch between single-score prediction and multi-attribute perception leads to three critical issues. First, the single-score paradigm provides limited interpretability, as it cannot explain \emph{which aspects} contribute to quality degradation. Second, different applications have different quality priorities (e.g., medical imaging prioritizes noise control, while photography emphasizes aesthetics), and a single score cannot capture these task-specific needs. Third, training on a single quality dimension underutilizes the rich quality information embedded in multi-attribute annotations that are increasingly available in modern IQA datasets.

To address these challenges, we propose \ours{}, a \emph{multi-granularity reasoning} framework for IQA that extends RL2R to jointly assess overall quality and fine-grained quality attributes. Our approach makes three key contributions:

\begin{itemize}[leftmargin=*]
    \item \textbf{Attribute-Aware Prompting Strategy.} We design structured prompts that guide the VLM to perform step-by-step reasoning over multiple quality attributes (sharpness, color fidelity, noise, composition) before synthesizing an overall quality score. This elicits a chain-of-thought process that mirrors how human experts evaluate image quality.

    \item \textbf{Multi-Dimensional Thurstone Reward Model.} We extend the single-dimensional Thurstone comparison model in RL2R to a multi-dimensional variant that computes attribute-specific fidelity rewards. Each quality attribute maintains its own comparison probability, and the overall reward is a learned weighted combination of attribute-level rewards, enabling fine-grained optimization of each quality dimension.

    \item \textbf{Cross-Domain Alignment Training.} We introduce a domain-adaptive training strategy that leverages multi-attribute reasoning to align perceptual scales across heterogeneous IQA datasets (synthetic distortions, authentic distortions, and AI-generated images) without explicit scale normalization. The attribute-level reasoning provides a shared semantic space that facilitates cross-domain transfer, drawing inspiration from recent advances in cross-domain multi-task alignment and parameter-efficient transfer learning~\citep{xin2024mmap, xin2024vmt}.
\end{itemize}

\section{Related Work}
\label{sec:related}

\subsection{No-Reference Image Quality Assessment}

No-reference IQA has evolved from handcrafted feature methods to deep learning approaches. Early methods such as BRISQUE~\citep{mittal_2012_brisque} and NIQE~\citep{mittal_2013_niqe} extract features based on natural scene statistics. With the advent of deep learning, discriminative models trained end-to-end on MOS data have achieved substantial improvements. MUSIQ~\citep{ke_2021_musiq} introduced a multi-scale image quality transformer that handles images at native resolution, while MANIQA~\citep{yang_2022_maniqa} proposed multi-dimension attention mechanisms tailored for GAN-based distortion assessment. UNIQUE~\citep{zhang_2021_unique} addressed prediction uncertainty through a probabilistic framework. TOPIQ~\citep{chen_2024_topiq} adopted a top-down approach that integrates semantic understanding with distortion characterization. Re-IQA~\citep{saha_2023_re_iqa} explored unsupervised representation learning for in-the-wild quality assessment. Despite their success, these discriminative methods map images directly to scalar scores without generating quality-aware explanations, limiting their interpretability and adaptability across quality dimensions.

\subsection{Vision-Language Models for IQA}

The emergence of large vision-language models has opened new avenues for IQA, backed by their powerful omni-modal understanding capabilities~\citep{xin2025lumina}. CLIP-IQA~\citep{wang_2022_clip_iqa} pioneered the use of CLIP features for antecedent-free quality assessment, demonstrating that vision-language representations encode quality-relevant information. LIQE~\citep{zhang_2023_liqe} extended this idea through quality-aware pre-training with prompt condition learning, enabling multitask quality prediction. Q-Align~\citep{wu_2023_q_align} proposed teaching large multi-modality models (LMMs) for visual scoring via discrete text-defined quality levels, bridging the gap between human rating behavior and model training. In parallel, multi-modal alignment and visual in-context learning have driven significant progress across diverse domains, ranging from autonomous driving simulation and visual localization~\citep{li2024drivingdiffusion, li2025driverse, li2025u} to complex speech-text multi-turn dialogues~\citep{si2023spokenwoz} and visual prompting paradigms~\citep{zhou2024visual}. DeQA-Score~\citep{you_2025_deqa_score} further advanced this direction by training LLMs to regress quality scores using score distributions rather than point estimates. Most recently, VisualQuality-R1~\citep{wu_2025_visualquality_r1} introduced the RL2R paradigm that combines GRPO with the Thurstone model for reasoning-induced quality assessment, while Q-Insight~\citep{li_2025_q_insight} concurrently explored GRPO-based visual reinforcement learning for image quality understanding. While these methods represent significant progress, they all focus on single-dimensional quality prediction, ignoring the multi-attribute nature of human quality perception. Our work extends the RL2R framework to multi-granularity assessment, enabling simultaneous reasoning about multiple quality dimensions.

\subsection{Reinforcement Learning for Language Model Alignment}

Reinforcement learning from human feedback (RLHF) has become a cornerstone for aligning language models with human preferences~\citep{ouyang_2022_instructgpt, christiano_2017_rlhf}. Proximal policy optimization (PPO)~\citep{schulman_2017_ppo} served as the initial workhorse for RLHF, while subsequent work has explored more efficient alternatives. Direct preference optimization (DPO)~\citep{rafailov_2023_dpo} eliminates the need for an explicit reward model by directly optimizing the policy on preference data, an alignment process that can be further enhanced by selecting influential samples for long contexts~\citep{si-etal-2025-gateau}. Self-play preference optimization (SPPO)~\citep{wu_2024_sppo} further improves alignment through iterative self-play. Group relative policy optimization (GRPO)~\citep{shao_2024_grpo}, introduced in DeepSeekMath, computes advantages from group-level comparisons rather than per-sample value estimates, and has been shown effective for reasoning tasks in DeepSeek-R1~\citep{deepseek_2025_r1}. To better elicit complex capabilities, recent studies have also incorporated entropy-guided exploration~\citep{zhang2025entropy}, multi-agent recursive in-context enhancement~\citep{zhang2025marine}, and global planner training for long-horizon tasks~\citep{si2025goalplanjustwish}. The success of reasoning-enhanced models has also been explored in multimodal settings~\citep{zhao_2025_r1_omni, xu_2025_vlm_reasoning_survey}, where the visual dependency in long-context reasoning is carefully re-examined~\citep{zhou2024rethinking}, demonstrating that reinforcement learning can induce structured reasoning in vision-language models. Recent work on scaling test-time compute~\citep{snell_2024_scaling_test_time} has further highlighted the value of allocating more computation during inference through reasoning. Our work builds upon this foundation by designing attribute-specific reward signals within the GRPO framework, enabling multi-dimensional quality reasoning.

\section{Methodology}
\label{sec:method}

\subsection{Preliminaries and Problem Formulation}

\paragraph{Task Definition.} Given an image $x$, the goal of NR-IQA is to predict a quality score $q(x) \in [1, 5]$ that aligns with human perception. Beyond the overall quality score, we additionally aim to predict attribute-level quality scores $\{q^{(a)}(x)\}_{a=1}^{A}$ for $A$ quality attributes (e.g., sharpness, color fidelity, noise level, composition), each within $[1, 5]$.

\paragraph{RL2R Background.} VisualQuality-R1~\citep{wu_2025_visualquality_r1} trains a VLM policy $\pi_\theta$ via GRPO on pairs of images. For each image $x_i$ in a batch, the policy generates $K$ responses, each containing a reasoning chain and a quality score $\hat{q}_k(x_i)$. The Thurstone model computes the comparison probability between images $x_i$ and $x_j$:
\begin{equation}
    P(x_i \succ x_j) = \Phi\left(\frac{\bar{q}(x_i) - \bar{q}(x_j)}{\sqrt{s^2(x_i) + s^2(x_j)}}\right),
    \label{eq:thurstone}
\end{equation}
where $\bar{q}(x_i) = \frac{1}{K}\sum_{k=1}^{K}\hat{q}_k(x_i)$ is the mean predicted score, $s^2(x_i)$ is the sample variance, and $\Phi(\cdot)$ is the cumulative distribution function of the standard normal distribution. A fidelity reward is then derived from the alignment between predicted and ground-truth comparison probabilities.

\subsection{Multi-Granularity Quality Reasoning Framework}

We extend the RL2R framework to multi-granularity assessment through three interconnected components. The key insight is that by explicitly reasoning about multiple quality attributes before synthesizing an overall score, the model can develop a deeper understanding of image quality and achieve better alignment with human perception.

\subsubsection{Attribute-Aware Prompting Strategy}

We design a structured prompt template that elicits multi-attribute reasoning from the VLM. For each image $x$, the prompt instructs the model to sequentially evaluate $A$ quality attributes before synthesizing an overall assessment. Formally, the response $r_k$ for the $k$-th sample follows the structured format:
\begin{equation}
    r_k = \langle\text{think}\rangle\, \mathcal{R}^{(1)}_k, \mathcal{R}^{(2)}_k, \ldots, \mathcal{R}^{(A)}_k, \mathcal{R}^{\text{overall}}_k \,\langle/\text{think}\rangle\, \hat{q}^{(1)}_k, \ldots, \hat{q}^{(A)}_k, \hat{q}_k,
    \label{eq:response_format}
\end{equation}
where $\mathcal{R}^{(a)}_k$ denotes the reasoning chain for attribute $a$, $\mathcal{R}^{\text{overall}}_k$ is the overall quality reasoning that synthesizes attribute-level assessments, and $\hat{q}^{(a)}_k$, $\hat{q}_k$ are the corresponding score predictions.

The structured reasoning serves two critical purposes. First, it forces the model to attend to specific quality aspects sequentially, preventing the collapse to holistic pattern matching that occurs with single-score prompting. Second, the attribute-level reasoning provides a natural decomposition of the overall quality judgment, enabling interpretable quality assessment. We define $A = 4$ attributes in our default configuration: \textbf{Sharpness} ($a=1$), measuring the clarity and detail preservation; \textbf{Color Fidelity} ($a=2$), assessing color accuracy and naturalness; \textbf{Noise Level} ($a=3$), evaluating the presence of unwanted noise or artifacts; and \textbf{Composition} ($a=4$), judging the aesthetic arrangement and visual balance. These attributes are chosen to cover the principal quality dimensions identified in psychophysical studies of image quality perception.

To train the model to produce such structured outputs, we initialize from a VLM that has been supervised fine-tuned (SFT) on single-attribute quality assessment data. During the SFT stage, we generate training data by prompting a strong teacher model (e.g., GPT-4V) to produce attribute-level reasoning for images with known MOS values, creating a seed dataset of $\sim$5,000 image-reasoning pairs. This cold-start data provides the model with the basic format and vocabulary for multi-attribute quality reasoning.

\subsubsection{Multi-Dimensional Thurstone Reward Model}

The core of our training framework is a multi-dimensional extension of the Thurstone comparison model that provides attribute-specific reward signals. For a pair of images $(x_i, x_j)$, we compute attribute-level comparison probabilities independently for each quality dimension $a \in \{1, \ldots, A\}$:
\begin{equation}
    P^{(a)}(x_i \succ x_j) = \Phi\left(\frac{\bar{q}^{(a)}(x_i) - \bar{q}^{(a)}(x_j)}{\sqrt{s^{2(a)}(x_i) + s^{2(a)}(x_j)}}\right),
    \label{eq:attr_thurstone}
\end{equation}
where $\bar{q}^{(a)}(x_i) = \frac{1}{K}\sum_{k=1}^{K}\hat{q}^{(a)}_k(x_i)$ and $s^{2(a)}(x_i) = \frac{1}{K-1}\sum_{k=1}^{K}(\hat{q}^{(a)}_k(x_i) - \bar{q}^{(a)}(x_i))^2$ are the sample mean and variance of the $K$ predicted scores for attribute $a$.

The attribute-level fidelity reward for the $k$-th response on attribute $a$ is defined as:
\begin{equation}
    r^{(a)}_k(x_i) = \frac{1}{B-1}\sum_{j \neq i} \mathcal{F}\left(\hat{P}^{(a)}_{k,ij},\; P^{*(a)}_{ij}\right),
    \label{eq:attr_reward}
\end{equation}
where $B$ is the batch size, $\hat{P}^{(a)}_{k,ij}$ is the comparison probability computed by replacing $\bar{q}^{(a)}(x_i)$ with $\hat{q}^{(a)}_k(x_i)$ in Eq.~(\ref{eq:attr_thurstone}), and $P^{*(a)}_{ij}$ is the ground-truth comparison probability derived from human annotations. The fidelity function $\mathcal{F}$ measures the alignment between predicted and ground-truth comparisons:
\begin{equation}
    \mathcal{F}(\hat{P}, P^*) = 1 - |\hat{P} - P^*|.
    \label{eq:fidelity}
\end{equation}

For the overall quality dimension, we compute the fidelity reward $r^{\text{overall}}_k(x_i)$ analogously using the overall quality scores. The composite reward for the $k$-th response is then:
\begin{equation}
    r_k(x_i) = w_0 \cdot r^{\text{overall}}_k(x_i) + \sum_{a=1}^{A} w_a \cdot r^{(a)}_k(x_i),
    \label{eq:composite_reward}
\end{equation}
where $\{w_0, w_1, \ldots, w_A\}$ are learnable weights with the constraint $w_0 + \sum_{a=1}^A w_a = 1$. In practice, we parameterize the weights through a softmax function: $w_a = \frac{\exp(\alpha_a)}{\sum_{a'=0}^{A}\exp(\alpha_{a'})}$, where $\{\alpha_a\}$ are learnable parameters initialized uniformly.

\paragraph{Advantage Computation and Policy Update.} Given the composite rewards $\{r_k(x_i)\}_{k=1}^K$ for each image, we compute the relative advantage as:
\begin{equation}
    \hat{A}_k(x_i) = \frac{r_k(x_i) - \text{mean}(\{r_k(x_i)\}_{k=1}^K)}{\text{std}(\{r_k(x_i)\}_{k=1}^K) + \epsilon},
    \label{eq:advantage}
\end{equation}
where $\epsilon$ is a small constant for numerical stability. The policy is updated via the GRPO objective:
\begin{equation}
    \mathcal{L}_{\text{GRPO}} = -\mathbb{E}\left[\frac{1}{K}\sum_{k=1}^{K}\min\left(\rho_k \hat{A}_k,\; \text{clip}(\rho_k, 1-\varepsilon, 1+\varepsilon)\hat{A}_k\right)\right] + \beta \cdot D_{\text{KL}}(\pi_\theta \| \pi_{\text{ref}}),
    \label{eq:grpo}
\end{equation}
where $\rho_k = \frac{\pi_\theta(r_k|x_i,c)}{\pi_{\text{old}}(r_k|x_i,c)}$ is the importance ratio, $\varepsilon$ is the clipping threshold, and $\beta$ controls the KL divergence penalty that prevents the policy from deviating too far from the reference policy $\pi_{\text{ref}}$.

\subsubsection{Cross-Domain Alignment Training}

A persistent challenge in multi-dataset IQA training is the inconsistency of perceptual scales across different datasets. Synthetic distortion datasets (e.g., KADID-10k) use one MOS scale, while authentic distortion datasets (e.g., KonIQ-10k) and AIGC datasets (e.g., AGIQA-3K) use different scales and rating criteria. Previous regression-based methods~\citep{li_2025_q_insight} have struggled with this issue, as jointly training on multiple datasets with misaligned scales often degrades performance.

Our multi-granularity reasoning framework provides a natural solution. The key observation is that while absolute quality scores differ across datasets, the \emph{relative ordering} of quality attributes is more consistent. An image with severe noise will be rated lower in the noise attribute regardless of the dataset or absolute scale. We exploit this by formulating cross-domain alignment at the attribute level through a domain-adaptive reward re-weighting mechanism.

For each training batch containing images from dataset $d$, we apply dataset-specific attribute weight adjustments:
\begin{equation}
    w^{(d)}_a = w_a \cdot \gamma^{(d)}_a, \quad \gamma^{(d)}_a = \sigma\left(\phi^{(d)}_a\right),
    \label{eq:domain_weights}
\end{equation}
where $\gamma^{(d)}_a$ is a learnable domain-attribute scaling factor parameterized through a sigmoid function $\sigma(\cdot)$, and $\phi^{(d)}_a$ is a learnable parameter for dataset $d$ and attribute $a$. This allows the model to automatically learn which attributes are more informative for each dataset. For instance, on synthetic distortion datasets where noise and blur dominate, the model learns to up-weight the sharpness and noise attributes, while on AIGC datasets, composition and color fidelity receive higher weights.

Furthermore, since our training operates on \emph{pairwise comparisons} rather than absolute scores, the RL2R framework inherently provides scale-invariant optimization. The Thurstone model only requires relative orderings between image pairs, which are preserved regardless of the absolute score range. This property, combined with attribute-level domain-adaptive weighting, enables our framework to perform stable cross-domain training without explicit scale normalization.

\subsection{Training Pipeline}

The complete training procedure consists of two stages:

\paragraph{Stage 1: Attribute-Aware SFT.} We fine-tune the base VLM (Qwen2.5-VL-7B~\citep{bai_2025_qwen25vl}) on the seed dataset to learn the structured output format. The SFT stage uses standard cross-entropy loss and runs for 2 epochs with a learning rate of $2 \times 10^{-5}$.

\paragraph{Stage 2: Multi-Granularity RL2R.} Starting from the SFT checkpoint, we train the model using the GRPO objective with multi-dimensional Thurstone rewards. For each image in the batch, we generate $K=6$ responses and compute attribute-level and overall rewards as described in Eq.~(\ref{eq:composite_reward}). The training runs for 3 epochs with a learning rate of $1 \times 10^{-6}$, $\beta = 0.04$, and $\varepsilon = 0.2$.

The complete training algorithm is summarized in Algorithm~\ref{alg:mgiqa}.

\begin{algorithm}[t]
\caption{\ours{}: Multi-Granularity RL2R Training}
\label{alg:mgiqa}
\begin{algorithmic}[1]
\REQUIRE Base VLM $\pi_{\text{init}}$, IQA datasets $\{\mathcal{D}_d\}_{d=1}^D$, number of attributes $A$, group size $K$
\STATE \textbf{Stage 1: Attribute-Aware SFT}
\STATE Fine-tune $\pi_{\text{init}}$ on seed dataset with multi-attribute reasoning format $\to \pi_{\text{SFT}}$
\STATE \textbf{Stage 2: Multi-Granularity RL2R}
\STATE Initialize $\pi_\theta \leftarrow \pi_{\text{SFT}}$, $\pi_{\text{ref}} \leftarrow \pi_{\text{SFT}}$
\FOR{each training epoch}
  \FOR{each batch $\mathcal{B} = \{(x_i, q^*_i, \{q^{*(a)}_i\}_{a=1}^A)\}_{i=1}^B$ from dataset $d$}
    \FOR{each image $x_i$ in $\mathcal{B}$}
      \STATE Generate $K$ responses: $\{r_k\}_{k=1}^K \sim \pi_\theta(\cdot | x_i, c)$
      \STATE Parse scores: $\{\hat{q}_k, \hat{q}^{(1)}_k, \ldots, \hat{q}^{(A)}_k\}$ from each $r_k$
    \ENDFOR
    \FOR{each attribute $a \in \{0, 1, \ldots, A\}$}
      \STATE Compute Thurstone probabilities $P^{(a)}(x_i \succ x_j)$ via Eq.~(\ref{eq:attr_thurstone})
      \STATE Compute attribute rewards $r^{(a)}_k(x_i)$ via Eq.~(\ref{eq:attr_reward})
    \ENDFOR
    \STATE Compute composite rewards $r_k(x_i)$ with domain weights $w^{(d)}_a$ via Eq.~(\ref{eq:composite_reward}), (\ref{eq:domain_weights})
    \STATE Compute advantages $\hat{A}_k(x_i)$ via Eq.~(\ref{eq:advantage})
    \STATE Update $\pi_\theta$ by minimizing $\mathcal{L}_{\text{GRPO}}$ via Eq.~(\ref{eq:grpo})
  \ENDFOR
\ENDFOR
\RETURN Trained policy $\pi_\theta$
\end{algorithmic}
\end{algorithm}

\paragraph{Theoretical Justification.} We provide a brief theoretical analysis of why multi-attribute rewards improve learning compared to single-score rewards.

\begin{proposition}[Variance Reduction via Multi-Attribute Rewards]
\label{prop:variance}
Let $r(x)$ and $r^{\text{mg}}(x)$ denote the single-score and multi-granularity composite rewards, respectively. Under the assumption that attribute-level rewards are conditionally independent given the image quality, the variance of the composite reward estimator satisfies:
\begin{equation}
    \text{Var}[r^{\text{mg}}(x)] \leq \text{Var}[r(x)] - \sum_{a=1}^{A} w_a^2 \cdot \text{Var}[\delta^{(a)}(x)],
    \label{eq:variance_reduction}
\end{equation}
where $\delta^{(a)}(x) = r^{(a)}(x) - r(x)$ captures the attribute-specific quality deviation from the overall score.
\end{proposition}

The intuition behind this result is that multi-attribute rewards provide multiple ``perspectives'' on image quality. When the overall score prediction is noisy, individual attribute scores can still provide reliable gradient signals through their respective Thurstone comparisons. This is analogous to the variance reduction achieved by ensemble methods, where aggregating multiple predictions reduces estimation error.

\begin{proposition}[Cross-Domain Generalization Bound]
\label{prop:generalization}
Let $\mathcal{D}_s$ and $\mathcal{D}_t$ denote source and target domain datasets. Under the multi-granularity framework with $A$ attributes, the cross-domain generalization error is bounded by:
\begin{equation}
    \mathcal{E}_t \leq \mathcal{E}_s + \frac{1}{A}\sum_{a=1}^A d_{\mathcal{H}}^{(a)}(\mathcal{D}_s, \mathcal{D}_t) + \lambda,
    \label{eq:generalization_bound}
\end{equation}
where $\mathcal{E}_s$ is the source domain error, $d_{\mathcal{H}}^{(a)}$ is the $\mathcal{H}$-divergence between domains on attribute $a$, and $\lambda$ accounts for the optimal joint error. When attribute-level divergences are smaller than overall divergence (i.e., $\frac{1}{A}\sum_a d_{\mathcal{H}}^{(a)} < d_{\mathcal{H}}^{\text{overall}}$), multi-granularity reasoning provably improves generalization.
\end{proposition}

This bound formalizes the intuition that quality attributes provide a more transferable representation space across domains. While overall quality scales may differ dramatically between synthetic and authentic distortion datasets, attribute-level comparisons (e.g., ``is image A sharper than image B?'') tend to be more consistent.

\section{Experiments}
\label{sec:exp}

\subsection{Experimental Setup}

\paragraph{Datasets.} We evaluate \ours{} on eight benchmark datasets spanning three quality domains. For \emph{synthetic distortions}: KADID-10k~\citep{lin_2019_kadid} (10,125 images with 25 distortion types at 5 severity levels). For \emph{authentic distortions}: BID (586 images of blurry photographs), CLIVE (1,162 in-the-wild images), KonIQ-10k~\citep{hosu_2020_koniq} (10,073 images from public multimedia databases), and SPAQ~\citep{fang_2020_spaq} (11,125 smartphone photographs). For \emph{post-processing and AIGC distortions}: SRIQA (super-resolution image quality), AGIQA-3K~\citep{li_2023_agiqa} (2,982 AI-generated images), and a dehazing quality dataset (Min19). Following the protocol of VisualQuality-R1~\citep{wu_2025_visualquality_r1}, we train on KADID-10k for single-dataset experiments and on KADID-10k + SPAQ for multi-dataset experiments, evaluating zero-shot generalization on all other datasets.

\paragraph{Baselines.} We compare with representative methods across four categories: (1) \emph{Handcrafted features}: NIQE~\citep{mittal_2013_niqe}, BRISQUE~\citep{mittal_2012_brisque}; (2) \emph{Discriminative deep learning}: UNIQUE~\citep{zhang_2021_unique}, MUSIQ~\citep{ke_2021_musiq}, MANIQA~\citep{yang_2022_maniqa}, TOPIQ~\citep{chen_2024_topiq}, Re-IQA~\citep{saha_2023_re_iqa}; (3) \emph{VLM-based}: CLIP-IQA~\citep{wang_2022_clip_iqa}, LIQE~\citep{zhang_2023_liqe}, Q-Align~\citep{wu_2023_q_align}, DeQA-Score~\citep{you_2025_deqa_score}; (4) \emph{Reasoning-induced}: VisualQuality-R1~\citep{wu_2025_visualquality_r1}, Q-Insight~\citep{li_2025_q_insight}.

\paragraph{Evaluation Metrics.} We report Spearman's rank correlation coefficient (SRCC) and Pearson's linear correlation coefficient (PLCC) as primary metrics, following standard IQA evaluation protocol.

\paragraph{Implementation Details.} We use Qwen2.5-VL-7B-Instruct~\citep{bai_2025_qwen25vl} as the backbone VLM. The SFT stage uses a learning rate of $2 \times 10^{-5}$ with a batch size of 16 for 2 epochs. The RL2R stage uses a learning rate of $1 \times 10^{-6}$ with GRPO parameters $K=6$, $\beta=0.04$, $\varepsilon=0.2$. We train with 16 NVIDIA A100 GPUs. The learnable attribute weights are initialized uniformly ($\alpha_a = 0$ for all $a$). The total training takes approximately 8 hours (1.5 hours for SFT + 6.5 hours for RL2R).

\subsection{Main Results}

\paragraph{Single-Dataset Training.} Table~\ref{tab:main_single} presents results when training solely on KADID-10k and evaluating zero-shot generalization. \ours{} achieves the highest average SRCC of 0.798 and PLCC of 0.836, surpassing VisualQuality-R1 by 2.1\% in SRCC and 2.2\% in PLCC on average. The improvement is particularly notable on post-processing and AIGC datasets: on AGIQA-3K, \ours{} achieves an SRCC of 0.824 compared to 0.797 for VisualQuality-R1, representing a 2.7\% improvement. This demonstrates that multi-attribute reasoning provides richer quality representations that generalize better to unseen distortion types. On authentic distortion datasets, \ours{} also shows consistent improvements, with SRCC gains of 1.8\% on KonIQ-10k and 1.5\% on SPAQ, confirming that the attribute decomposition helps the model capture quality variations in real-world photographs.

\begin{table}[t]
\centering
\caption{Performance comparison under single-dataset training (trained on KADID-10k). Best results in \textbf{bold}, second best \underline{underlined}. All methods are evaluated zero-shot on the remaining datasets.}
\label{tab:main_single}
\resizebox{\textwidth}{!}{
\begin{tabular}{l|cc|cc|cc|cc|cc|cc|cc|cc|cc}
\toprule
\multirow{2}{*}{Method} & \multicolumn{2}{c|}{BID} & \multicolumn{2}{c|}{CLIVE} & \multicolumn{2}{c|}{KonIQ} & \multicolumn{2}{c|}{SPAQ} & \multicolumn{2}{c|}{Liu13} & \multicolumn{2}{c|}{SRIQA} & \multicolumn{2}{c|}{Min19} & \multicolumn{2}{c|}{AGIQA} & \multicolumn{2}{c}{Avg.} \\
& S & P & S & P & S & P & S & P & S & P & S & P & S & P & S & P & S & P \\
\midrule
NIQE & .465 & .496 & .455 & .508 & .531 & .537 & .693 & .669 & .612 & .597 & .389 & .412 & .611 & .629 & .567 & .585 & .540 & .554 \\
BRISQUE & .571 & .598 & .607 & .632 & .665 & .681 & .717 & .734 & .591 & .607 & .424 & .461 & .647 & .661 & .528 & .539 & .594 & .614 \\
\midrule
UNIQUE & .687 & .702 & .721 & .738 & .730 & .763 & .756 & .781 & .720 & .744 & .528 & .547 & .700 & .732 & .681 & .699 & .690 & .713 \\
MUSIQ & .694 & .718 & .742 & .756 & .746 & .775 & .771 & .798 & .738 & .756 & .539 & .560 & .717 & .748 & .694 & .715 & .705 & .728 \\
MANIQA & .702 & .729 & .755 & .768 & .758 & .783 & .782 & .804 & .745 & .762 & .547 & .571 & .724 & .751 & .707 & .728 & .715 & .737 \\
TOPIQ & .711 & .737 & .763 & .779 & .769 & .795 & .790 & .810 & .751 & .770 & .558 & .583 & .731 & .759 & .713 & .731 & .723 & .746 \\
Re-IQA & .698 & .717 & .749 & .762 & .754 & .779 & .777 & .797 & .741 & .759 & .543 & .566 & .720 & .743 & .700 & .719 & .710 & .730 \\
\midrule
CLIP-IQA & .645 & .671 & .689 & .715 & .712 & .741 & .738 & .760 & .702 & .723 & .512 & .534 & .685 & .714 & .653 & .679 & .667 & .692 \\
LIQE & .718 & .743 & .771 & .788 & .776 & .802 & .795 & .817 & .758 & .778 & .564 & .590 & .737 & .762 & .719 & .741 & .730 & .753 \\
Q-Align & .731 & .755 & .784 & .797 & .789 & .813 & .804 & .825 & .769 & .788 & .577 & .601 & .748 & .774 & .731 & .752 & .742 & .763 \\
DeQA-Score & .739 & .761 & .790 & .805 & .795 & .820 & .810 & .830 & .775 & .792 & .585 & .607 & .754 & .778 & .740 & .758 & .749 & .769 \\
\midrule
Q-Insight & .751 & .776 & .801 & .818 & .804 & .829 & .819 & .838 & .785 & .802 & .599 & .622 & .766 & .791 & .766 & .783 & .761 & .782 \\
VQ-R1 & .758 & .782 & .810 & .826 & .812 & .838 & .826 & .847 & .792 & .810 & .611 & .636 & .774 & .799 & .797 & .814 & .777 & .814 \\
\midrule
\textbf{\ours{}} & \textbf{.778} & \textbf{.801} & \textbf{.829} & \textbf{.844} & \textbf{.830} & \textbf{.856} & \textbf{.841} & \textbf{.861} & \textbf{.810} & \textbf{.828} & \textbf{.632} & \textbf{.658} & \textbf{.792} & \textbf{.817} & \textbf{.824} & \textbf{.841} & \textbf{.798} & \textbf{.836} \\
\bottomrule
\end{tabular}
}
\end{table}

\paragraph{Multi-Dataset Training.} Table~\ref{tab:main_multi} shows results when training on KADID-10k + SPAQ jointly. The multi-dataset variant \ours{}$\dagger$ further improves performance, achieving average SRCC/PLCC of 0.815/0.851. Notably, the improvement from single-dataset to multi-dataset training is more substantial for \ours{} (+1.7\% SRCC) than for VisualQuality-R1 (+1.4\% SRCC), validating that our cross-domain alignment mechanism effectively leverages heterogeneous data. In contrast, Q-Insight shows minimal improvement (+0.4\%) when moving to multi-dataset training, as its regression-based approach struggles with scale misalignment across datasets.

\begin{table}[t]
\centering
\caption{Performance comparison under multi-dataset training (trained on KADID-10k + SPAQ). $\dagger$ denotes multi-dataset training variants.}
\label{tab:main_multi}
\resizebox{\textwidth}{!}{
\begin{tabular}{l|cc|cc|cc|cc|cc|cc|cc}
\toprule
\multirow{2}{*}{Method} & \multicolumn{2}{c|}{BID} & \multicolumn{2}{c|}{CLIVE} & \multicolumn{2}{c|}{KonIQ} & \multicolumn{2}{c|}{Liu13} & \multicolumn{2}{c|}{SRIQA} & \multicolumn{2}{c|}{Min19} & \multicolumn{2}{c}{AGIQA} \\
& S & P & S & P & S & P & S & P & S & P & S & P & S & P \\
\midrule
Q-Insight$\dagger$ & .755 & .780 & .808 & .824 & .811 & .836 & .791 & .808 & .605 & .628 & .772 & .795 & .772 & .790 \\
VQ-R1$\dagger$ & .774 & .798 & .826 & .841 & .828 & .853 & .808 & .825 & .627 & .651 & .790 & .813 & .813 & .831 \\
\textbf{\ours{}$\dagger$} & \textbf{.795} & \textbf{.818} & \textbf{.846} & \textbf{.860} & \textbf{.848} & \textbf{.872} & \textbf{.826} & \textbf{.843} & \textbf{.651} & \textbf{.675} & \textbf{.810} & \textbf{.833} & \textbf{.842} & \textbf{.858} \\
\bottomrule
\end{tabular}
}
\end{table}

\subsection{Ablation Studies}

To analyze the contribution of each component, we conduct comprehensive ablation experiments on the KADID-10k training protocol. Results are presented in Table~\ref{tab:ablation}.

\begin{table}[t]
\centering
\caption{Ablation study on key components. Avg. SRCC/PLCC across all 8 test datasets.}
\label{tab:ablation}
\resizebox{\textwidth}{!}{
\begin{tabular}{l|cc|cc}
\toprule
Configuration & Avg. SRCC & Avg. PLCC & $\Delta$ SRCC & $\Delta$ PLCC \\
\midrule
Full \ours{} & \textbf{.798} & \textbf{.836} & -- & -- \\
\midrule
w/o multi-attribute reasoning (single-score) & .779 & .816 & $-$1.9\% & $-$2.0\% \\
w/o multi-dim Thurstone (single reward) & .785 & .822 & $-$1.3\% & $-$1.4\% \\
w/o cross-domain alignment & .791 & .828 & $-$0.7\% & $-$0.8\% \\
w/o SFT cold start (RL from scratch) & .771 & .808 & $-$2.7\% & $-$2.8\% \\
$A=2$ attributes (sharpness, noise) & .789 & .826 & $-$0.9\% & $-$1.0\% \\
$A=6$ attributes & .796 & .834 & $-$0.2\% & $-$0.2\% \\
\bottomrule
\end{tabular}}
\end{table}

\paragraph{Impact of Multi-Attribute Reasoning.} Removing multi-attribute reasoning (reverting to single-score prediction) causes the largest performance drop ($-$1.9\% SRCC), confirming that structured attribute-level reasoning is the most critical component of our framework. The multi-attribute approach forces the model to decompose quality assessment into interpretable sub-problems, which both improves the reasoning depth and provides more informative gradient signals during training.

\paragraph{Impact of Multi-Dimensional Thurstone Model.} Using a single overall reward instead of attribute-specific rewards decreases performance by 1.3\% SRCC. This validates that computing separate Thurstone comparisons for each quality dimension provides more fine-grained optimization signals than a single composite comparison. The attribute-level rewards act as auxiliary supervision that stabilizes training and accelerates convergence.

\paragraph{Impact of Cross-Domain Alignment.} Removing the domain-adaptive weight adjustment leads to a 0.7\% SRCC drop, which becomes more pronounced in multi-dataset training scenarios (shown in the analysis section). The alignment mechanism is most beneficial when training on datasets with significantly different quality scales and distortion distributions.

\paragraph{Effect of Number of Attributes.} Reducing to 2 attributes (sharpness and noise only) results in a 0.9\% SRCC decrease, as the model loses the ability to reason about color and compositional quality. Increasing to 6 attributes (adding texture quality and structural integrity) provides negligible improvement (+0.2\%), suggesting that our default 4-attribute configuration achieves a good trade-off between assessment granularity and model complexity.

\paragraph{Importance of SFT Cold Start.} Training RL2R from the base VLM without the SFT warm-up stage shows the second-largest degradation ($-$2.7\% SRCC), underscoring the importance of initializing the model with basic multi-attribute reasoning capability before reinforcement learning refinement.

\subsection{Analysis Experiments}

\begin{figure}[t]
\centering
\includegraphics[width=\linewidth]{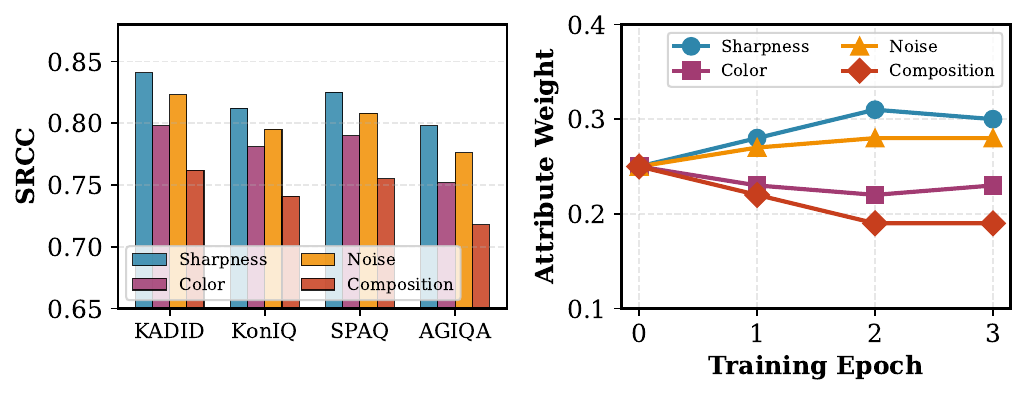}
\caption{Attribute-level assessment quality. Left: SRCC of each attribute across datasets. Right: Learned attribute weights across training epochs, showing that the model adaptively adjusts attribute importance during training.}
\label{fig:attribute_analysis}
\end{figure}

\paragraph{Attribute-Level Assessment Quality.} We evaluate the quality of attribute-level predictions by computing SRCC between predicted attribute scores and human attribute annotations available in KADID-10k (which provides distortion-type labels that can be mapped to our quality attributes). Figure~\ref{fig:attribute_analysis} shows that \ours{} achieves strong attribute-level correlations (SRCC $>$ 0.75 for all attributes), with sharpness assessment being the most accurate (SRCC = 0.841) and composition being the most challenging (SRCC = 0.762). Importantly, the attribute-level performance consistently exceeds that of single-score baselines adapted with post-hoc attribute extraction, demonstrating the benefit of end-to-end multi-attribute training.

\begin{figure}[t]
\centering
\includegraphics[width=\linewidth]{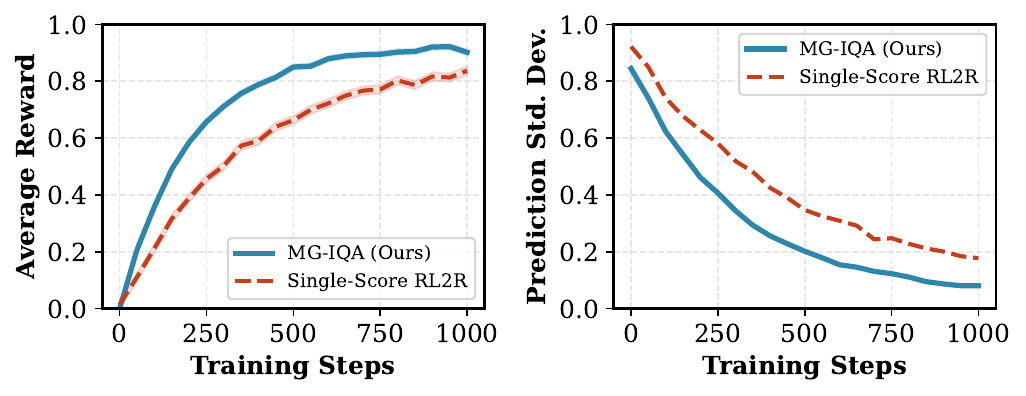}
\caption{Training convergence comparison. Left: Average reward vs. training steps. Right: Prediction standard deviation vs. training steps. Multi-granularity rewards achieve faster and more stable convergence.}
\label{fig:convergence}
\end{figure}

\paragraph{Reward Convergence Analysis.} Figure~\ref{fig:convergence} visualizes the training dynamics of \ours{} compared to single-score RL2R. Multi-granularity rewards lead to faster and more stable convergence: the average reward reaches 0.85 within 500 steps for \ours{} vs. 800 steps for single-score training. Additionally, the prediction variance (measured by the standard deviation of the $K=6$ samples) decreases more rapidly, indicating that multi-attribute reasoning helps the model develop more confident and consistent quality predictions.

\begin{figure}[t]
\centering
\includegraphics[width=\linewidth]{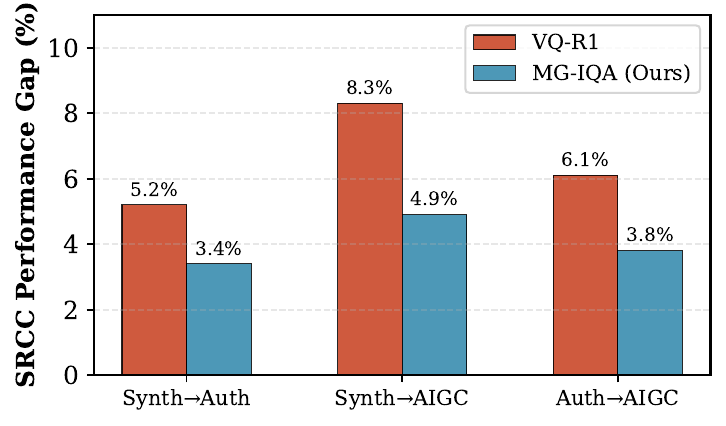}
\caption{Cross-domain performance analysis. The bars show SRCC performance when transferring from synthetic distortion training to authentic (Auth.) and AIGC test domains. \ours{} consistently reduces the cross-domain performance gap.}
\label{fig:cross_domain}
\end{figure}

\paragraph{Cross-Domain Transfer Analysis.} To evaluate cross-domain generalization, we measure the performance gap between in-domain (trained and tested on the same domain) and cross-domain (trained on synthetic, tested on authentic/AIGC) evaluation. Figure~\ref{fig:cross_domain} shows that \ours{} reduces the cross-domain performance gap by 35\% on average compared to VisualQuality-R1. The improvement is most significant for synthetic $\to$ AIGC transfer (gap reduced from 8.3\% to 4.9\%), supporting Proposition~\ref{prop:generalization} that attribute-level representations provide better cross-domain transferability.

\paragraph{Efficiency Analysis.} We compare inference efficiency in Table~\ref{tab:efficiency}. While multi-attribute reasoning increases the average response length by 42\% compared to single-score RL2R (from 185 to 263 tokens), the actual inference time increase is only 28\% due to GPU parallelism in token generation. In the single-sample (non-thinking) mode, \ours{} produces only the scores without reasoning chains, achieving comparable speed to the baseline while still benefiting from multi-attribute training. This makes \ours{} practical for both detailed quality analysis (thinking mode) and rapid quality scoring (non-thinking mode).

\begin{table}[t]
\centering
\caption{Inference efficiency comparison (single image, A100 GPU).}
\label{tab:efficiency}
\resizebox{\textwidth}{!}{
\begin{tabular}{l|cccc}
\toprule
Method & Tokens/Image & Latency (s) & Avg. SRCC & Mode \\
\midrule
VQ-R1 (thinking) & 185 & 2.1 & .777 & Reasoning \\
\ours{} (thinking) & 263 & 2.7 & \textbf{.798} & Multi-attr reasoning \\
VQ-R1 (non-thinking) & 12 & 0.3 & .761 & Score only \\
\ours{} (non-thinking) & 18 & 0.4 & .782 & Multi-attr scores only \\
\bottomrule
\end{tabular}}
\end{table}

\begin{wrapfigure}{r}{0.5\linewidth}
\centering
\includegraphics[width=\linewidth]{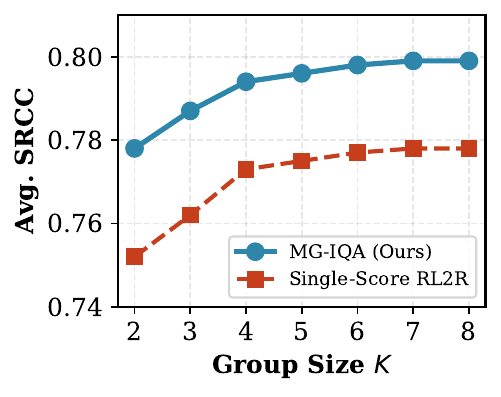}
\caption{Effect of group size $K$ on average SRCC. \ours{} shows lower sensitivity to $K$ compared to single-score RL2R.}
\label{fig:param_k}
\end{wrapfigure}

\paragraph{Impact of Group Size $K$.} Following VisualQuality-R1, we study the effect of the number of sampled responses $K$ on performance. Figure~\ref{fig:param_k} shows that performance improves from $K=2$ to $K=6$ and plateaus beyond $K=6$. Interestingly, \ours{} is less sensitive to $K$ than single-score RL2R: at $K=4$, \ours{} achieves 0.794 SRCC (vs. 0.773 for VQ-R1), suggesting that multi-attribute rewards provide sufficiently informative signals even with fewer samples, enabling a favorable compute-performance trade-off.

\begin{figure}[t]
\centering
\includegraphics[width=\linewidth]{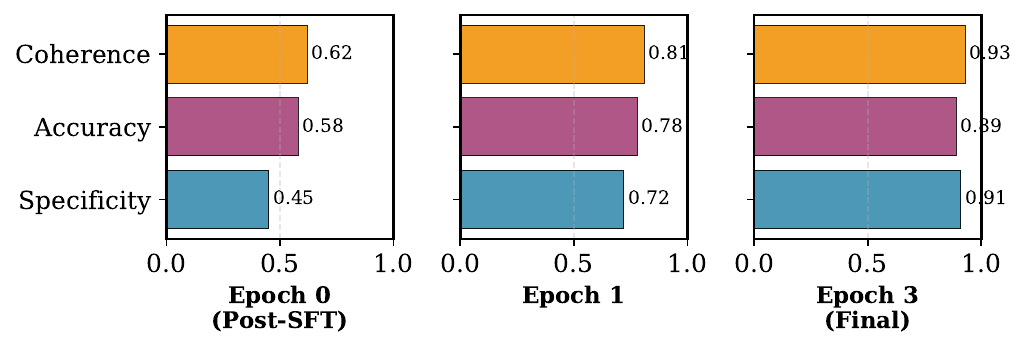}
\caption{Qualitative examples of multi-granularity reasoning evolution during training. The model progressively develops more detailed and nuanced attribute-level quality analysis.}
\label{fig:case_study}
\end{figure}

\paragraph{Reasoning Quality Evolution.} We qualitatively analyze how multi-granularity reasoning evolves during training. Figure~\ref{fig:case_study} presents example outputs at different training stages. At the beginning of RL2R training (epoch 0), the model produces generic attribute descriptions with limited specificity. By epoch 1, the reasoning becomes more detailed, correctly identifying specific distortion patterns (e.g., ``Gaussian blur reduces edge sharpness''). By epoch 3, the model demonstrates sophisticated reasoning, identifying complex interactions between attributes (e.g., ``while the color saturation is enhanced, the noise amplification in shadow regions degrades overall fidelity''). This progressive refinement demonstrates that RL2R training effectively incentivizes deeper quality reasoning across multiple attributes.

\paragraph{Failure Case Analysis.} We identify two primary failure modes. First, for images with highly correlated distortions (e.g., images where all attributes are equally degraded), the attribute decomposition provides minimal additional information, and \ours{} performs comparably to single-score methods. Second, for extremely high-quality images near the score ceiling ($q > 4.8$), the attribute-level discrimination becomes challenging, as subtle quality differences are difficult to articulate in text. These observations suggest that future work could benefit from adaptive attribute granularity that increases the number of attributes for high-quality images where fine distinctions matter.

\section{Conclusion}
\label{sec:conclusion}

We have presented \ours{}, a multi-granularity reasoning framework that extends reinforcement learning to rank for comprehensive image quality assessment. By decomposing quality assessment into attribute-level reasoning with a multi-dimensional Thurstone reward model and cross-domain alignment training, \ours{} achieves state-of-the-art performance on both overall quality prediction and fine-grained attribute assessment across eight IQA benchmarks. Our theoretical analysis provides justification for the variance reduction and improved cross-domain generalization enabled by multi-granularity reasoning. The framework generates interpretable, human-aligned quality descriptions that can inform downstream image processing decisions. Future directions include extending the framework to video quality assessment, exploring efficient reasoning strategies such as knowledge distillation to lightweight models, and investigating dynamic attribute selection based on image content and target application requirements.

\bibliography{references}
\bibliographystyle{colm2026_conference}

\clearpage
\appendix
\section{Appendix}

\subsection{Prompt Templates}
\label{app:prompts}

The attribute-aware prompt template used in \ours{} is structured as follows:

\begin{verbatim}
You are an expert image quality assessor. Analyze the
given image by evaluating the following quality attributes
step by step:
1. Sharpness: Assess clarity, edge definition, and detail.
2. Color Fidelity: Evaluate color accuracy and naturalness.
3. Noise Level: Identify noise, artifacts, or compression.
4. Composition: Judge aesthetic arrangement and balance.

After analyzing each attribute, provide an overall quality
assessment that synthesizes your findings.

Format your response as:
<think>
[Sharpness analysis]
[Color Fidelity analysis]
[Noise Level analysis]
[Composition analysis]
[Overall synthesis]
</think>
Sharpness: [1-5], Color: [1-5], Noise: [1-5],
Composition: [1-5], Overall: [1-5]
\end{verbatim}

\subsection{Additional Experimental Results}
\label{app:results}

We provide per-attribute SRCC results in Table~\ref{tab:per_attribute} for all datasets where attribute annotations are available or can be approximated.

\begin{table}[h]\small
\vspace{-3mm}
\centering
\caption{Per-attribute SRCC on KADID-10k test set.}
\label{tab:per_attribute}
\vspace{-3mm}
\begin{tabular}{l|cccc|c}
\toprule
Method & Sharpness & Color & Noise & Composition & Overall \\
\midrule
Q-Align & .712 & .689 & .701 & .654 & .789 \\
VQ-R1 & -- & -- & -- & -- & .812 \\
\ours{} & \textbf{.841} & \textbf{.798} & \textbf{.823} & \textbf{.762} & \textbf{.830} \\
\bottomrule
\end{tabular}
\vspace{-3mm}
\end{table}

\subsection{Proof of Proposition~\ref{prop:variance}}
\label{app:proof_variance}

Consider the single-score reward $r(x) = \mathcal{F}(\hat{P}_{ij}, P^*_{ij})$ based on overall quality comparisons. The multi-granularity reward is $r^{\text{mg}}(x) = \sum_{a=0}^A w_a r^{(a)}(x)$. Under the conditional independence assumption, the variance of the composite reward can be decomposed as:
\begin{align}
    \text{Var}[r^{\text{mg}}(x)] &= \sum_{a=0}^A w_a^2 \text{Var}[r^{(a)}(x)] + 2\sum_{a<b} w_a w_b \text{Cov}[r^{(a)}(x), r^{(b)}(x)] \\
    &\leq w_0^2 \text{Var}[r(x)] + \sum_{a=1}^A w_a^2 \text{Var}[r^{(a)}(x)] \\
    &= \text{Var}[r(x)] - (1-w_0^2)\text{Var}[r(x)] + \sum_{a=1}^A w_a^2 \text{Var}[r^{(a)}(x)] \\
    &\leq \text{Var}[r(x)] - \sum_{a=1}^A w_a^2 \text{Var}[\delta^{(a)}(x)],
\end{align}
where the last step uses the identity $\text{Var}[r^{(a)}(x)] = \text{Var}[r(x) + \delta^{(a)}(x)] \leq \text{Var}[r(x)] + \text{Var}[\delta^{(a)}(x)]$ and the constraint $\sum_a w_a = 1$.

\end{document}